\pdfoutput=1

\documentclass[11pt]{article}

\usepackage[]{acl}

\usepackage{times}
\usepackage{latexsym}

\usepackage[T1]{fontenc}

\usepackage[utf8]{inputenc}

\usepackage{microtype}

\usepackage{dblfloatfix}

\usepackage{caption}

\usepackage[bottom,hang]{footmisc}
\usepackage{times,latexsym}
\usepackage{url}
\usepackage[T1]{fontenc}
\usepackage{enumitem}
\usepackage{fancybox}
\usepackage{calc}
\usepackage{amsmath}
\usepackage{stmaryrd}
\usepackage{graphicx}       %
\usepackage{babel}
\usepackage{color}
\usepackage{multirow}
\usepackage{todonotes}
\usepackage{soul}
\usepackage{booktabs}
\usepackage{adjustbox}
\usepackage{array}
\usepackage{tabularx}
\usepackage{paralist} %
\usepackage{bbding} %
\usepackage{arydshln}

\usepackage{comment}

\newcolumntype{R}[2]{%
    >{\adjustbox{angle=#1,lap=\width-(#2)}\bgroup}%
    l%
    <{\egroup}%
}
\newcommand*\rot{\multicolumn{1}{R{45}{1em}}}%

\newcommand\tab[1][0.7cm]{\hspace*{#1}}
\newcommand{\affa}{{$^{\dagger}$}}
\newcommand{\affb}{{$^{\ddagger}$}}
\newcommand{\affc}{{$^{\star}$}}
\newcommand{\affd}{{$^{\triangle}$}}
\newcommand{\affe}{{$^{\diamond}$}}

\newcommand{\babibm}{{bAbI 1.0}\xspace}

\newcommand{\pybabi}{{Dyna-bAbI}\xspace}

\newcommand{\pybabiurl}{\url{https://tiny.one/8wjxwd7z}} %

\newcommand{\writingdate}{November 11, 2021}
\newcommand{\questionset}{$\mathcal{Q}^{*}$\xspace}
\newcommand{\lingset}{$\mathcal{L}^{*}$\xspace}
\newcommand{\eventset}{$\mathcal{E}^{*}$\xspace}

\newcommand{\possquestionset}{$\mathcal{Q}$\xspace}
\newcommand{\posslingset}{$\mathcal{L}$\xspace}
\newcommand{\posseventset}{$\mathcal{E}$\xspace}

\newcommand{\suppcomp}{$f_{c}$\xspace}
\newcommand{\suppsize}{$\left|f\right|$\xspace}

\newcommand{\concat}[1]{\textit{concat(#1)}}
\newcommand{\mix}[1]{\textit{mix(#1)}}
\newcommand{\diverse}[1]{\textit{diverse(#1)}}

\newcommand{\concatnarg}{\textit{concat}\xspace}
\newcommand{\mixnarg}{\textit{mix}\xspace}
\newcommand{\diversenarg}{\textit{diverse}\xspace}
\newcommand{\injectnarg}{\textit{inject}\xspace}

\newcommand{\reasonpat}{support compositions\xspace}

\newcommand{\wherep}{\textit{where-P}\xspace}
\newcommand{\whereo}{\textit{where-O}\xspace}
\newcommand{\wherewaso}{\textit{where-was-O}\xspace}
\newcommand{\yesno}{\textit{yes-no}\xspace}
\newcommand{\listq}{\textit{list}\xspace}
\newcommand{\countq}{\textit{count}\xspace}

\usepackage[linesnumbered,ruled,vlined]{algorithm2e}

\usepackage{booktabs}

\SetCommentSty{mycommfont}

\SetKwInput{KwInput}{Input}                %
\SetKwInput{KwOutput}{Output}              %

\newcommand{\xhdr}[1]{\vspace{1mm}\noindent{{\bf #1.}}}

\colorlet{shadecolor}{gray!40}
\usepackage{xcolor,colortbl}

\title{Dyna-bAbI: unlocking bAbI's potential with dynamic synthetic benchmarking}

\author{Ronen Tamari\affa\thanks{Work done during an internship at the Allen Institute.} \tab Kyle Richardson\affc \tab Aviad Sar-Shalom\affd \tab Noam Kahlon\affa \\ \quad {\bf Nelson F. Liu}\affe \tab  {\bf Reut Tsarfaty}\affc\affb \tab {\bf Dafna Shahaf}\affa \\
  \affa The Hebrew University of Jerusalem  \tab \affc Allen Institute for AI \\
   \tab \affb Bar-Ilan University \tab \affd Tel-Aviv University \tab \affe Stanford University \\
  \texttt{\{ronent,dshahaf\}@cs.huji.ac.il},  \texttt{\{reutt,kyler\}@allenai.org} %
}

\begin{document}
\maketitle
\begin{abstract}

While neural language models often perform surprisingly well on natural language understanding (NLU) tasks, their strengths and limitations remain poorly understood. Controlled synthetic tasks are thus an increasingly important resource for diagnosing model behavior. In this work we focus on story understanding, a core competency for NLU systems. However, the main synthetic resource for story understanding, the bAbI benchmark, lacks such a systematic mechanism for controllable task generation. We develop \pybabi, a dynamic framework providing fine-grained control over task generation in bAbI. We demonstrate our ideas by constructing three new tasks requiring compositional generalization, an important evaluation setting absent from the original benchmark. We tested both special-purpose models developed for bAbI as well as state-of-the-art pre-trained methods, and found that while both approaches solve the original tasks (>99\% accuracy), neither approach succeeded in the compositional generalization setting, indicating the limitations of the original training data.
We explored ways to augment the original data, and found that though diversifying training data was far more useful than simply increasing dataset size, it was still insufficient for driving robust compositional generalization (with <70\% accuracy for complex compositions). Our results underscore the importance of highly controllable task generators for creating robust NLU systems through a virtuous cycle of model and data development.\footnote{Code and data will be made available at \pybabiurl}

\end{abstract}

\section{Introduction}
Considerable progress has been made recently in natural language understanding (NLU), driven largely by advances in model pre-training \cite{devlin-etal-2019-bert,2020t5}  and the development of large-scale NLU benchmarks across a wide range of tasks \cite{wang-etal-2018-glue,wang2019superglue,liang-etal-2020-xglue}.
\begin{figure}[t!]
\centering
\includegraphics[width=\columnwidth]{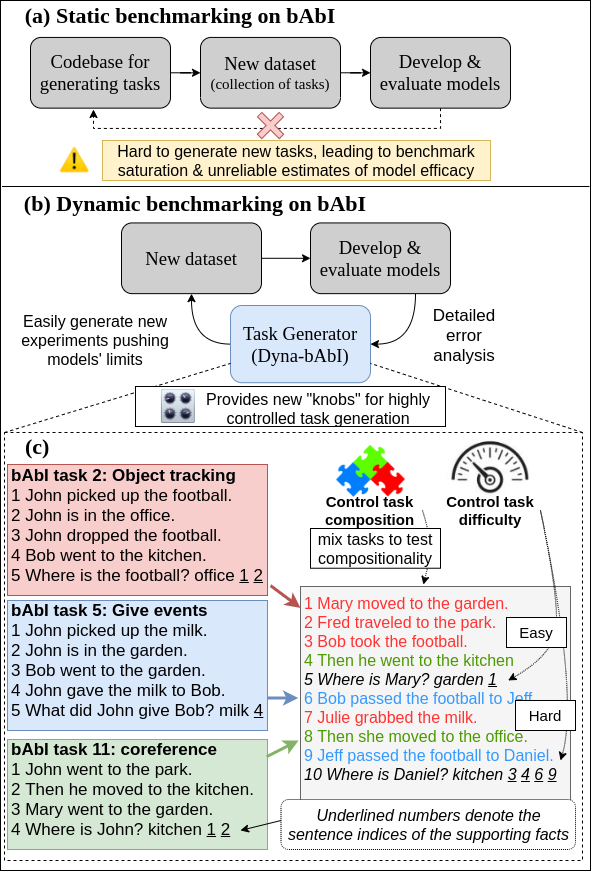}

\caption{\label{fig:sdb-overview}
(a) Low task configurability leads to static datasets, benchmark saturation \& unreliable model development. (b) We propose a dynamic benchmarking approach; developing models and tasks in a tight feedback loop using (c) \pybabi task generator. \pybabi provides fine-grained control over task structure, composition and difficulty, yielding challenging new test sets exposing limitations of state-of-the-art models.}
\end{figure}
Such successes, however, have coincided with the discovery of various shortcomings in existing human curated datasets, largely related to \emph{annotation artifacts} \citep{gururangan-etal-2018-annotation}, or systematic biases that create shortcuts that can inflate model performance and harm generalization. 

In order to overcome these issues, two avenues of research have recently gained traction: 1)~the development of \emph{dynamic benchmarks} \cite{potts-etal-2021-dynasent,kiela2021dynabench} where, in contrast to conventional \emph{static} benchmarks, evaluation and data collection are preformed in an agile manner and conducted interactively with humans and models in a rapidly evolving feedback loop and; 2)~renewed interest in  \emph{synthetic benchmarks} \cite{lake2018generalization,sinha-etal-2019-clutrr,clark2020transformers,ruis2020benchmark} that allow for absolute control over the data creation process in order to help understand the strengths and weaknesses of existing models on targeted tasks and language phenomena.

Story understanding is a particularly important domain for research on dynamic and synthetic benchmarks; it is a core competency for NLU systems~\citep{McClelland2020,dunietz-etal-2020-test}, but the scale and annotation detail required make human data collection prohibitively costly. However, the main synthetic resource for story understanding remains the bAbI task suite~\citep{babi2016}, which is saturated by models reaching near-perfect performance~\citep{liu2021small}, and further limited by exploitable biases in the data~\citep{kaushik-lipton-2018-much}. Despite its creators' initial intentions, bAbI has largely remained a static benchmark limited to a small subset of the tasks potentially possible to generate within the bAbI ``micro-world''. Accordingly, two natural questions arise: \textbf{(Q1)} \emph{is near-perfect model performance on the original bAbI tasks a reliable indicator of story understanding competence?}; \textbf{(Q2)} \emph{are there still interesting challenges to discover inside the broader bAbI task space that help identify weaknesses in current models and drive modeling innovation?}

To answer these questions, we employ a \emph{dynamic synthetic benchmarking} approach on bAbI, combining the benefits of the agile approach of recent dynamic benchmarks with the scale and control provided by synthetic datasets. As illustrated in Figure~\ref{fig:sdb-overview}, in dynamic synthetic benchmarks the data generator itself is designed for agile development, enabling experimentation with increasingly complex tasks and a wider range of linguistic phenomena. Constructing challenging tasks is a challenge in and of itself, requiring precise control over the reasoning patterns underlying each question. To meet these requirements, we developed a new task generator for bAbI called \pybabi\footnote{Implemented in Python for improved accessibility compared with the original Lua implementation (\url{https://github.com/facebookarchive/bAbI-tasks}).}.

Using \pybabi, we first devise new splits that systematically test \emph{compositional generalization} across tasks; as shown in Fig. \ref{fig:sdb-overview}c, we test models on novel combinations (line 10 question on right) of concepts  seen at training, like co-reference and object tracking (left). We find that training on the original bAbI tasks (hereafter: \babibm) is not sufficient for models to attain good compositional generalization. Though general purpose pre-trained models far outperform special-purpose (non-pre-trained) architectures developed for bAbI, they still suffer a 30-50\% drop in accuracy compared to the non-pre-trained models which suffer a 50-80\% drop. Both types attain near perfect performance on the original tasks, suggesting that \babibm is not challenging enough to differentiate between the two classes of models \textbf{(Q1)}. 

We next investigate how different enhancements of training data affect compositional generalization: (a) injecting more questions into \babibm, and (b) generating new, more diverse training samples. Compared to question injection, we find that diverse training data better facilitates compositional generalization, as well as being more data efficient. However, neither approach drives \emph{reliable} compositional generalization; a representative state-of-the-art (SOTA) model, T5~\citep{2020t5}, demonstrates a lack of robustness to novel combinations and also exhibits knowledge inconsistency, for example, correctly answering certain types of questions, but systematically failing to answer equivalent question paraphrases. These results suggest that there remain many important challenges within the broader bAbI task space \textbf{(Q2)}.

To sanity-check the quality of our new tests compared with \babibm, we employ the notion of \emph{concurrence} proposed by \citet{liu2021small}; concurrence is a measure of correlation between models' performance on a synthetic task and their performance on an existing, non-synthetic NLU benchmark. We find high concurrence between our new challenge tasks and the widely used SQuAD dataset~\citep{rajpurkar-etal-2016-squad}, in contrast to \babibm which achieved low concurrence. 

Giving the continued interest in using \babibm to evaluate new modeling approaches ~\citep{Banino2020MEMO,banino2021pondernet,schlag2021learning}, our new challenge splits and the \pybabi task generator contribute to more reliably guiding future efforts.
While we focused on bAbI, our results apply more generally, telling a cautionary tale about the limits of static synthetic datasets, and motivating the development of controllable task generators for dynamic synthetic benchmarking.

\section{Related Work}
\label{sec:related}
Our work brings together two promising areas of current research: dynamic benchmarking such as Dynabench~\citep{kiela2021dynabench} that address many  existing issues with static benchmarks~\citep{bowman-dahl-2021-will}, and synthetic benchmarking, which is widely used for high-precision and data-intensive problems such as relational and logical reasoning \cite{sinha-etal-2019-clutrr,clark2020transformers,betz2020critical}, robot planning \cite{banerjee2020transformers}, instruction following and language grounding \cite{long-etal-2016-simpler,lake2018generalization} among many others \cite{richardson2020probing,khot2021learning}. Most approaches to synthetic benchmarking focus on model development on a static benchmark, and are not designed to facilitate agile and highly controlled task space exploration, which is our focus here. The recent gSCAN dataset~\citep{ruis2020benchmark} and later extensions~\citep{qiu2021systematic,wu2021reascan}
 can be seen as an example of a synthetic benchmark ``going dynamic''. Our work differs in terms of target domain (story understanding as opposed to multi-modal language grounding), and we further focus attention on a more general research direction of intentional, a-priori design of NLU benchmarks for agile development.

We address the domain of story understanding as a particularly core (and data-intensive) capacity underlying language use~\citep{McClelland2020}, thought to require constructing and manipulating situation models of entities and their relations as they unfold throughout discourse~\citep{Zwaan2016,tamari-etal-2020-language}.  Procedural text datasets~\citep{dalvi-etal-2018-tracking,tandon-etal-2020-dataset} are closely related in that they provide detailed annotation of entities and state changes, and have mostly focused on relatively small and static benchmarks using human collected data. Overall, recent works identify a lack of benchmark tasks which systematically probe the situation models constructed by NLP systems processing discourse-level texts~\citep{sugawara-etal-2021-benchmarking}. 

The bAbI benchmark~\cite{babi2016} is seen as highly relevant in terms of objective (targeting situation modelling)~\citep{dunietz-etal-2020-test}, but has been viewed critically due to its constrained nature and exploitable artifacts~\citep{kaushik-lipton-2018-much}. Our work focuses on improving the evaluation in bAbI through compositional generalization, widely used across NLP to more rigorously probe model robustness~\citep{finegan-dollak-etal-2018-improving,keysers2020measuring,gontier2020measuring,yanaka-etal-2021-sygns}, but to our knowledge still not applied to story understanding or bAbI.

\section{Synthetic Dynamic Benchmarking on bAbI}

\subsection{\pybabi}
\label{ssec:pybabi}
What makes a synthetic benchmark \emph{dynamic}? We think of a dynamic synthetic benchmark as a highly controllable task generator, enabling rapid exploration of interesting areas of task space. The original \babibm simulator code does not readily facilitate such exploration; each of the \babibm tasks is generated by a hard-coded script which does not neatly expose ``dials'' to manipulate interesting generation aspects such as question difficulty or compositionality. 

Accordingly, we developed \pybabi, a Python-based version of the original simulator. \pybabi facilitates control of task generation through a configuration file, effectively abstracting away much of the underlying implementation complexity. The configuration file allows users to specify high-level task parameters such as the concepts set, passage length, and filtering conditions to mine for harder/rarer examples. We also modularized the code to facilitate adding new questions and other concepts more easily.

In this next sections we describe the underlying structure of the \babibm tasks, and how we combine them using \pybabi to create more complex compositional generalization tasks.

\subsection{bAbI task structure}
\label{ssec:task_structure}
A task in \babibm is a set of train, validation and test splits. Each split is a set of instances, where an instance is a tuple ($p,q,a$)=(\textit{passage, question, answer}). Passages are generated using a micro-world simulator, by sampling a valid sequence of world events from an event set \eventset and generating a linguistic description of them. By default, linguistic descriptions are generated by a simple sentence-level mapping from an event to a natural language sentence. For example, the event \texttt{move(john,park)} could be translated to ``John moved to the park.'' Some tasks also incorporate more complex linguistic mappings between events and sentences, such as co-reference: the event sequence (\texttt{move(john,park)}, \texttt{move(john,kitchen)}) could be mapped to ``John moved to the park. Then he went to the kitchen.'' 

We denote the set of possible linguistic mappings by \lingset. Finally, a valid question-answer pair ($q$,$a$) over $p$ is sampled from question set \questionset. In bAbI, splits are usually generated using a subset of possible events, linguistic constructs and questions; we denote these as \posseventset,\posslingset,\possquestionset, respectively. We can then define the \emph{concept set} of a specific split,  $\mathcal{C}=\mathcal{E}\cup\mathcal{L}\cup\mathcal{Q}$.
Instances also include a set of supporting facts ($f$), or the relevant lines from which $a$ can be derived (see Fig. \ref{fig:sdb-overview}). The support composition ($f_{c}$) is the set of events and linguistic constructs contained in $f$ (see examples in \S\ref{ssec:exp_2_err_analysis}), and is useful for characterizing compositionality performance (\S\ref{ssec:comp_gen_motivation}).

\subsection{Original \babibm tasks}

\begin{table}[t]
\small
\begin{tabular}{@{}c@{\hspace{1\tabcolsep}}c@{\hspace{1\tabcolsep}}c@{\hspace{1\tabcolsep}}c@{\hspace{1\tabcolsep}}c@{}}
\toprule
Task & Events                                                        & \begin{tabular}[c]{@{}c@{}}Linguistic \\ Constructs\end{tabular} & \multicolumn{1}{c}{Questions} & \begin{tabular}[c]{@{}c@{}}Avg. sents. \& supp. \\ facts per story\end{tabular} \\ \midrule
1    & MOVE                                                          & -                                                                & where-P                       & 6, 1                                                                            \\
2    & \begin{tabular}[c]{@{}c@{}}MOVE, \\ POSS\end{tabular}         & -                                                                & where-O                       & 15.52, 2                                                                        \\
3    & \begin{tabular}[c]{@{}c@{}}MOVE, \\ POSS\end{tabular}         & -                                                                & where-was-O                   & 51.9, 3                                                                         \\
5    & \begin{tabular}[c]{@{}c@{}}MOVE, \\ GIVE,\\ POSS\end{tabular} & -                                                                & give-qs                       & 20.1, 1                                                                         \\
6    & MOVE                                                          & -                                                                & yes-no                        & 6.27, 1                                                                         \\
7    & \begin{tabular}[c]{@{}c@{}}MOVE, \\ GIVE,\\ POSS\end{tabular} & -                                                                & counting                      & 8.67, 2.33                                                                      \\
8    & \begin{tabular}[c]{@{}c@{}}MOVE, \\ POSS\end{tabular}         & -                                                                & list                          & 8.75, 1.94                                                                      \\
9    & MOVE                                                          & NEGATE                                                           & yes-no                        & 6, 1                                                                            \\
10   & MOVE                                                          & INDEF                                                            & yes-no                        & 6, 1                                                                            \\
11   & MOVE                                                          & CO-REF                                                           & where-P                       & 6, 2                                                                            \\
12   & MOVE                                                          & CONJ.                                                            & where-P                       & 6, 1                                                                            \\
13   & MOVE                                                          & \begin{tabular}[c]{@{}c@{}}CONJ., \\ CO-REF\end{tabular}         & where-P                       & 6, 2                                                                            \\ \bottomrule
\end{tabular}
\caption{\label{tab:babi_20_tasks} Subset of 12 \babibm tasks considered here. Each task is characterized by the possible events, linguistic constructs and questions that can occur in instances. POSS (possession) is short for GRAB and DROP events. Statistics based on training sets. A large space of task configurations remains unexplored.}
\end{table}

Our focus here is on a particular subset of 12 \babibm tasks evaluating aspects of story understanding. Table \ref{tab:babi_20_tasks} summarizes them, detailing \posseventset,\posslingset,\possquestionset for each task. For \posslingset, we list only complex constructs beyond the default event-sentence mapping (which is present in every task). See appendix \ref{ssec:ext_task_details} for additional details on task construction. Not all of the story understanding tasks are considered. For example, tasks 14 and 20 address time reasoning and agent motivations, and we leave their integration for future work.

\subsection{Compositional generalization on bAbI}
\label{ssec:comp_gen_motivation}

As can be seen in Table \ref{tab:babi_20_tasks}, many possible task configurations are not covered by the original benchmark; which directions should be explored? We focus on out-of-distribution (OOD) robustness, which is increasingly seen as a vital evaluation criteria across AI/NLP research~\citep{shanahan2020artificial,hendrycks-etal-2020-pretrained}. In particular, we target the OOD capacity for \emph{compositional generalization}; the ability to systematically generalize to test inputs containing novel combinations of more basic elements seen at training time~\citep{partee1995lexical,lake2017building}. For example, a model that has learned basic object tracking and co-reference \emph{separately} (tasks 2 and 11, see Fig. \ref{fig:sdb-overview}c) could be expected to solve tasks requiring a \emph{mixture} of both object tracking and co-reference (Fig. \ref{fig:sdb-overview}c, line 10 question on right side). Compositional tasks are absent from \babibm which features only IID test sets (independent, identically distributed).\footnote{\citet{babi2016} noted that transfer learning was an important goal out of the original work's scope.}

\xhdr{Compositional task generation} To create compositional generalization tasks in practice, we create training (and validation) splits composed of $M$ sub-tasks with concept sets $\left\{ \mathcal{C}_{\text{train}}^{i}\right\} _{i=1}^{M}$, and a test set $\mathcal{C}_{\text{test}}$ such that $\mathcal{C}_{\text{test}}\neq\mathcal{C}_{\text{train}}^{i}\forall i$, but $\mathcal{C}_{\text{test}}=\bigcup_{i=1}^{M}\mathcal{C}_{\text{train}}^{i}$. In other words, each training sub-task can be thought of focusing on a particular subset of test concepts, so models are exposed to all test concepts at training time, but not to all combinations of them~\citep{yanaka-etal-2021-sygns}. 

\xhdr{Task difficulty} We hypothesize that support composition (\suppcomp) and supporting fact set size (\suppsize) are main factors underlying a particular instance's difficulty, and especially \emph{novel} support compositions not seen at training time. Additionally, the difference between train and test splits results in potentially harder distractors, as test-time distractors appear in novel contexts.

Our notions of concept and support composition resemble atoms and compounds in DBCA, a related study on compositionality~\citep{keysers2020measuring}. While DBCA enables automatic creation of compositional train and test splits, we opt here for a more human-interpretable representation that allows more precise manual control of the combinations of concepts a model is exposed to at train and test time.

\xhdr{Quality comparison vs. \babibm tasks} Intuitively, good synthetic datasets help drive the development of better modelling approaches. Our new compositional tasks might be harder than \babibm, but how do we know whether they are a more useful target? To provide a preliminary answer to this question, we adopt the notion of \emph{concurrence} as a quality measure~\citep{liu2021small}. Two benchmarks are said to have high concurrence when they rank a set of modelling approaches similarly. Concurrence offers a way to formalize the intuition above, as high concurrence between a synthetic and natural language benchmark suggests that the synthetic benchmark could have driven similar innovations. We follow the setup of \citet{liu2021small} using SQuAD for the natural language benchmark.\footnote{\citet{liu2021small} consider a set of 20 modelling approaches used on SQuAD, including 10 pre-trained and 10 non-pre-trained methods.} Notably, \babibm achieved very low concurrence with SQuAD; for example, pre-training consistently yields large gains on SQuAD, but on \babibm, both pre-trained and non-pre-trained models achieve perfect performance on many tasks. The low concurrence thus suggests that \babibm may be an unreliable benchmark for model development, and highlights the importance of improving its quality.

\label{sec:sdb_babi}

\section{Experiments}
\label{sec:experiments}
With the controllable task generation afforded by \pybabi, we can now create datasets probing deeper story understanding capabilities of models.

We present two main experiments targeting the following questions:
\begin{compactitem}
    \item Exp. 1: (q1.a) What role does model architecture play in the capacity for compositional generalization? (q1.b) What is the concurrence of our compositional tasks with real datasets, compared with \babibm? 
    \item Exp. 2: (q2) How do training data quantity and diversity affect compositional generalization?
\end{compactitem}
\subsection*{Data} 

\begin{table}[!t]
\small
\begin{tabular}{@{}llllc@{}}
\toprule
Split        & Type  & \begin{tabular}[c]{@{}l@{}}Avg. \\ length\end{tabular} & Size    & \multicolumn{1}{l}{\begin{tabular}[c]{@{}l@{}}Avg. supp.\\ fact set size\end{tabular}} \\ \midrule
concat(T2)   & Train & 10.76                                                  & 18,000  & 2                                                                                      \\ \cdashline{1-5}
concat(T7)   & Train & 13.5                                                   & 63,000  & 1.68                                                                                   \\
inject(T7)   & Train & 23.25                                                  & 190,158 & 1.42                                                                                   \\
diverse(T7)  & Train & 20                                                     & 17,000  & 2.17                                                                                   \\ \cdashline{1-5}
concat(T12)  & Train & 10.8                                                   & 108,000 & 1.42                                                                                   \\
inject(T12)  & Train & 15.97                                                  & 368,831 & 1.28                                                                                   \\
diverse(T12) & Train & 20                                                     & 24,772  & 2.45                                                                                   \\ \midrule
mix(T2)      & Test  & 13.25                                                     & 1,000   & 2.05                                                                                       \\ \cdashline{1-5}
mix(T7)      & Test  & 20                                                     & 3,000   & 2.50                                                                                   \\ \cdashline{1-5}
mix(T12)     & Test  & 20                                                     & 6,000   & 3.70                                                                                   \\ \cdashline{1-5} \bottomrule
\end{tabular}
\caption{\label{tab:datasets} Splits used for our experiments. All except the original data (\concatnarg) are created with \pybabi.}
\end{table}

For our experiments we created 4 kinds of splits over three subsets of \babibm tasks, summarized in Table \ref{tab:datasets}. We denote a subset of tasks $T$, and consider $T_2=\left\{ 2,11\right\} $, $T_7=\left\{1,2,3,5,11,..,13\right\}$, and $T_{12}=\left\{1,2,3,5,...,13\right\}$.
\begin{compactitem}
    \item \concatnarg splits are simply concatenations of the official data for the tasks $T$. We considered the larger version where each task consists of 9,000/1,000 training/development examples; e.g., \concat{$T_2$} consists of 18,000 training examples and 2,000 development examples.
    \item \injectnarg splits enrich the \concatnarg data as follows: for each question in the original data, we supplement it with all possible additional questions of the specified types. In this work, the supplement question types were \wherep and \whereo (to provide location information of objects and agents).
    \item \diversenarg splits use rejection sampling to generate more diverse samples, such that the number of supporting facts per question is roughly uniform across all sub-task instances for a given question type. Without rejection sampling, most generated questions would be trivial (e.g., 1-2 supporting facts). Compositionality is retained by holding out certain combinations. In particular, at training time, complex linguistic constructs (e.g., co-reference) are only seen with MOVE events.
    \item \mixnarg are test splits generated using rejection sampling like \diversenarg, and consist of instances which may feature events, linguistic constructs and questions from any of the considered tasks. As a result, questions in \mixnarg splits require novel/more complex reasoning patterns compared to those seen at training time.
\end{compactitem}

See appendix \ref{ssec:ext_task_details} for examples and extended details on task generation.

\subsection{Exp. 1: Can training on \babibm facilitate compositional generalization?}

\begin{table*}[]
\small
\begin{tabular}{@{}llllllllll@{}}
\toprule
Name        & Train       & Test  & \multicolumn{5}{c}{Evaluation accuracy}  & \multicolumn{2}{l}{SQuAD Concurrence} \\ \midrule
            &             &             & EntNet & STM   & BiDAF & Roberta & T5    & $\rho$               & $\tau$               \\
2-task IID  & concat(T2)  & concat(T2)  & 98.95  & 99.85 & 100   & 100     & 99.85 & {[}-0.35,0.08{]}  & {[}-0.35,-0.19{]} \\
2-task OOD  & concat(T2)  & mix(T2)     & \textcolor{red}{\textbf{72.0}}     & \textcolor{red}{\textbf{67.6}}  & 97.2  & 98.7    & 98.1  & 0.48              & 0.51              \\ \cdashline{1-10}
7-task IID  & concat(T7)  & concat(T7)  & 96.8   & 99.4  & 99.98 & 99.98   & 99.8  & {[}-0.4,0.08{]}   & {[}-0.35,0.03{]}  \\
7-task OOD  & concat(T7)  & mix(T7)     & \textcolor{red}{\textbf{22.2}}   & \textcolor{red}{\textbf{26.7}}  & \textcolor{red}{\textbf{30.5}}  & \textcolor{red}{\textbf{57.7}}    & \textcolor{red}{\textbf{62.66}} & 0.92              & 0.78              \\ \cdashline{1-10}
12-task IID & concat(T12) & concat(T12) & 96.19  &   99.34    & --     & --       & 99.54 & --                 & --                 \\
12-task OOD & concat(T12) & mix(T12)    & \textcolor{red}{\textbf{31.97}}  &   \textcolor{red}{\textbf{35.65}}    & --     & --       & \textcolor{red}{\textbf{67.4}}  & --                 & --                 \\ \bottomrule
\end{tabular}
\caption{\label{tab:exp_1_results} Experiment 1. OOD evaluation exposes large differences between pre-trained and non-pre-trained models, and also achieves high concurrence with the SQuAD benchmark. We report [min,max] concurrence for \babibm.}
\end{table*}

For this experiment, we compared models on $T_2$ and $T_7$, since they allow for a direct conversion to an extractive QA format,\footnote{Tasks 6-10 require generative QA, for answering \yesno, \countq and \listq questions.} thus enabling us to use the same concurrence measurement framework of \citet{liu2021small}.

\xhdr{Models} We considered 3 classes of models: 
\begin{compactitem}
    \item  Non-pre-trained specialized architectures for \babibm including EntNet~\citep{entnet2017} and STM~\citep{le2020self}, the latter being current SOTA on \babibm\footnote{As of \writingdate.}.
    \item  Non-pretrained general-purpose QA methods, such as BiDAF~\citep{seo2016bidirectional}.
    \item General purpose pre-trained approaches including RoBERTa~\citep{liu2020roberta} and T5 (base)~\citep{2020t5}.
\end{compactitem}

The last two categories are comprised of the 20 models evaluated in \citet{liu2021small}, with the addition of T5 to the last group. For implementation details, see appendix \ref{ssec:impl}.

\subsubsection*{Results \& Analysis} 
\label{ssec:4_1_analysis}
Experiment results are summarized in Table \ref{tab:exp_1_results}. All models perform well in IID settings, but performance drops considerably in OOD settings, including for the SOTA STM model. Pre-trained models fare better on the OOD splits, but still suffer large drops for the harder 7 and 12-task OOD splits.

\xhdr{Architecture alone is not a significant compositionality driver (q1.a)} The large OOD performance gap between pre-trained and non-pre-trained models indicates that pre-training plays a much greater role than specialized architectures for QA performance, adding to similar findings in other NLP domains~\citep{hendrycks-etal-2020-pretrained}. The results raise questions about special purpose relational reasoning architectures that continue to be developed today: the poor OOD performance suggests that such models may not be fulfilling their intended design. Either way, we believe these results underscore the importance of rigorous evaluation to verify that modelling motivations are borne out in practice~\citep{aina-etal-2019-entity}. 

\xhdr{Compositionality increases concurrence (q1.b)} As can be seen in the Fig. \ref{fig:concurrence} plots\footnote{See appendix \ref{ssec:ext_concurrence} for full numeric results.}, increasing compositionality is correlated with increased concurrence. In particular, the 7-task OOD split yields high concurrence with the SQuAD benchmark, comparable to other \emph{natural} language as well as purpose-built synthetic datasets considered in \citet{liu2021small}, which feature $r,\tau$ (Pearson and Kendall correlation functions, resp.) in the ranges $\left[0.87,0.99\right]$ and $\left[0.77,0.94\right]$, respectively. Our results extend the findings of \citet{liu2021small}; they demonstrated the \emph{existence} of high concurrence synthetic benchmarks, we additionally suggest a guiding principle for how to \emph{create} them (compositional generalization).

\begin{figure}
\centering
\includegraphics[width=\columnwidth]{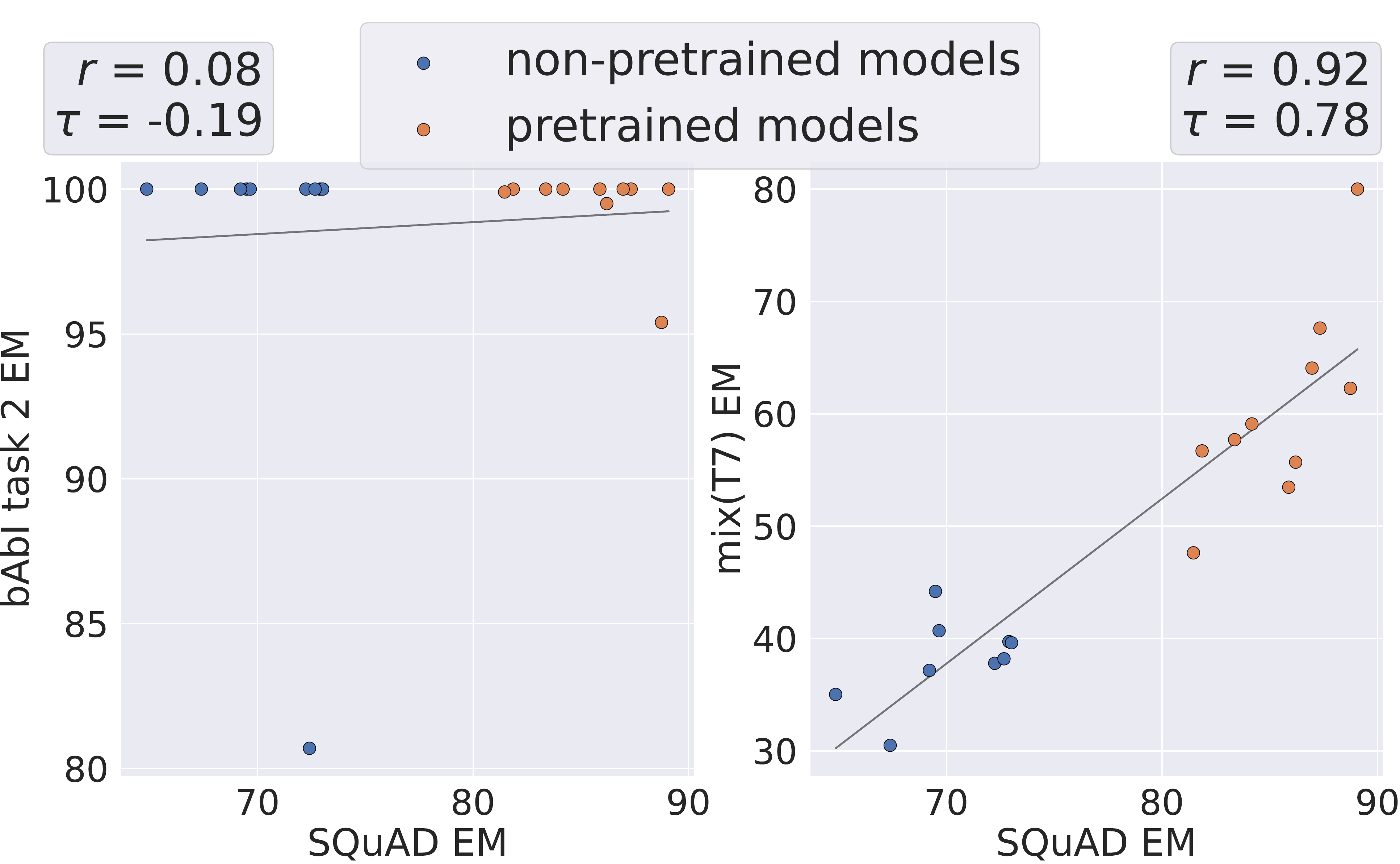}

\caption{\label{fig:concurrence} SQuAD concurrence plots for bAbI task 2 (reproduced from \citet{liu2021small} with permission) and \mix{$T_{7}$}. bAbI task 2 has the highest SQuAD concurrence of all $T_7$ tasks, yet is still significantly lower than \mix{$T_{7}$}, highlighting the relevance of compositional evaluation.}
\end{figure}

\subsection{Exp. 2: enriching \babibm training data} 
\label{ssec:exp_2}
The results above suggest that the bAbI data in their current form may not be rich enough to drive compositional generalization.\footnote{An alternate hypothesis is that certain patterns may be too hard for models to learn; we confirm this is not the case by using the inoculation methodology of \citet{liu-etal-2019-inoculation}, see details in Appendix \ref{ssec:ext_inoc}.} In this experiment we probe this  question, enriching the training data to better understand its impact on compositional generalization. In particular, we investigate two approaches to enriching the training data while maintaining the compositionality evaluation, corresponding to the \injectnarg and \diversenarg splits.

We focus on pre-trained models, as they significantly out-performed non-pre-trained methods. We use T5 as a representative since its generative abilities make it straightforward to apply also to $T_{12}$ (unlike the extractive methods which were applicable only to $T_7$).

\begin{table}[!t]
\small
\begin{tabular}{@{}l@{\hspace{1.2\tabcolsep}}l@{\hspace{1.2\tabcolsep}}rrrr@{}}
\toprule
Train        & Test      & \multicolumn{4}{l}{\begin{tabular}[c]{@{}l@{}}Evaluation accuracy /\\  \# supporting facts\end{tabular}} \\ \midrule
             &           & 1                        & 2                        & 3+                        & Total                   \\
inject(T7)   & concat(T7)  & 99.83                    & 100                      & 93.35                    & 99.05                   \\ 
inject(T7)   & mix(T7)   & 89.82                    & \textcolor{red}{\textbf{80.55}}                    & \textcolor{red}{\textbf{64.16}}                    & \textcolor{red}{\textbf{71.57}}                   \\ \cdashline{1-6}
diverse(T7)  & concat(T7)  & 99.58                    & 100                      & 78.36                    & 96.94                   \\
diverse(T7)  & mix(T7)   & 100                      & 98.44                    & 93.84                    & 95.8                    \\ \midrule
inject(T12)  & concat(T12) & 99.94                    & 99.97                    & 91.91                    & 99.35                   \\
inject(T12)  & mix(T12)  & 92.45                    & \textcolor{red}{\textbf{85.29}}                    & \textcolor{red}{\textbf{67.67}}                    & \textcolor{red}{\textbf{72.2}}                    \\ \cdashline{1-6}
diverse(T12) & concat(T12) & 99.75                    & 98.73                    & 76.81                    & 97.73                   \\
diverse(T12) & mix(T12)  & 99.01                    & 96.29                    & \textcolor{red}{\textbf{81.24}}                    & \textcolor{red}{\textbf{84.82}}                   \\ \bottomrule
\end{tabular}
\caption{\label{tab:exp_2_results} Enriching the training data. Injecting knowledge to the original bAbI tasks doesn't substantially improve compositionality. Sampling more structurally diverse instances yields more significant improvements, though is still limited, especially for more complex compositions. 
}
\end{table}

\xhdr{Injecting supplementary questions} One hypothesis for the poor performance of models on the \mixnarg splits could be that the original bAbI tasks do not provide enough supervision for models to learn the basic event semantics. For example, tasks 5 and 7 are the only \babibm tasks featuring the GIVE event, and neither includes any questions about the location of participants. However, test-time compositional questions may require models to infer that the participants in a GIVE event share the same location (e.g., line 10 question in Fig. \ref{fig:sdb-overview}c). Error analysis shows that such implicit inferences are indeed challenging for models trained on the \concatnarg splits (see appendix \ref{ssec:ext_err_analysis_give} for details). Perhaps the \injectnarg splits supplementing the original tasks with questions providing relevant information will improve compositionality performance? Table \ref{tab:exp_2_results} displays the result of this experiment; performance in the \mixnarg setting is improved only marginally, even though the amount of training data increases 3-fold (Table \ref{tab:datasets}).

\xhdr{Sampling structurally diverse training data} As shown in Table \ref{tab:datasets}, though \injectnarg splits significantly increase dataset size, their diversity remains low: most questions require only one or two supporting facts. Therefore, we next enrich training data through sampling more structurally diverse samples. This method is known to improve data efficiency for both compositional generalization as well as IID settings~\citep{oren2021finding}. As can be seen in Table \ref{tab:exp_2_results}, training on the \diversenarg splits yields a more significant improvement; similar to the findings of \citet{oren2021finding}, sampling more diverse training data leads to greater generalization as well as much improved data efficiency.\footnote{The relatively low performance of \diversenarg trained models in the ``3+'' column for \concatnarg splits is predominantly due to length discrepancies at train and test time: \concatnarg contains some very long stories which are challenging for the model trained on the uniform length and shorter \diversenarg stories.} However, as the error analysis of the next section shows, performance on compositional generalization is still fundamentally limited.

\subsubsection*{Discussion and error analysis} 
\label{ssec:exp_2_err_analysis}

\begin{figure*}[t!]
\centering
\includegraphics[width=1\linewidth]{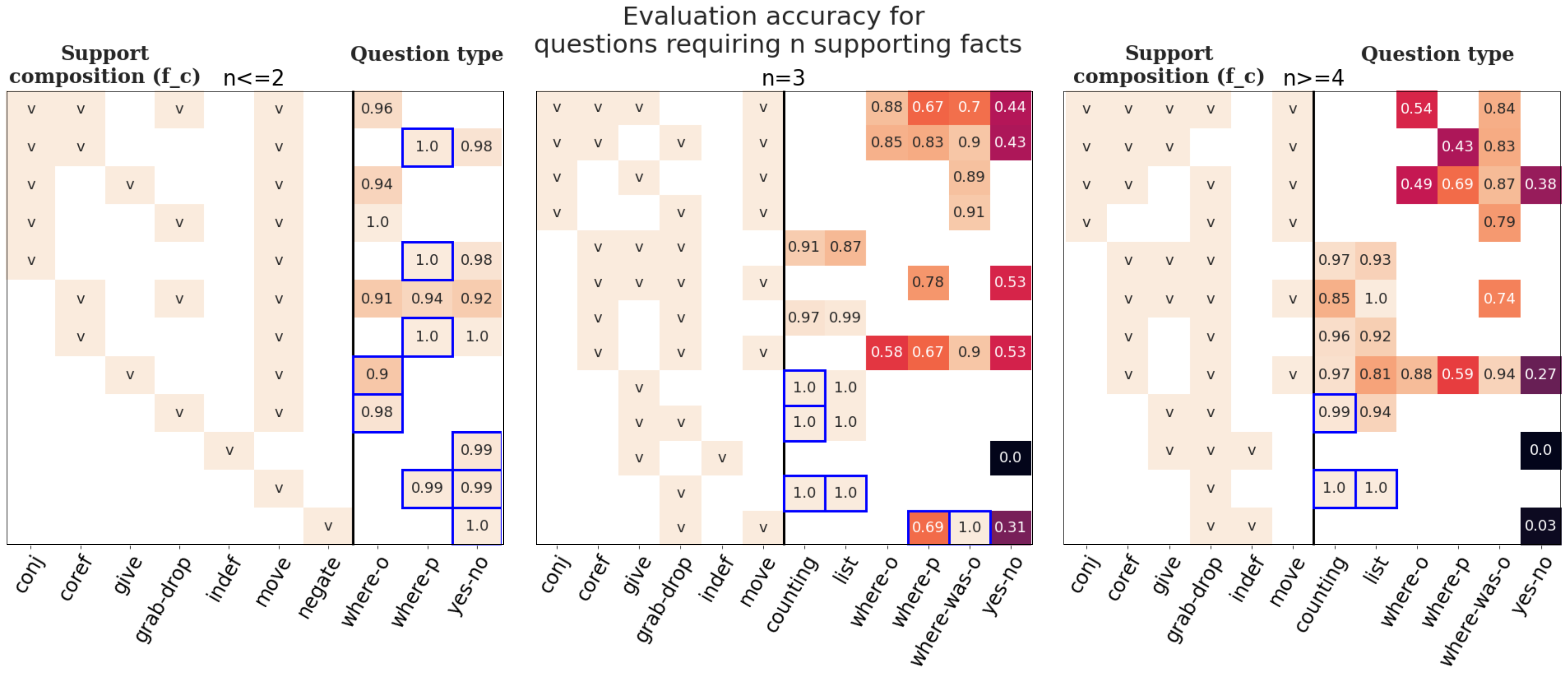}
\caption{\label{fig:heatmaps} Error analysis on \mix{$T_{12}$} for T5 trained on \diverse{$T_{12}$} data.  Performance on support compositions seen at training time (blue frames) is generally high, but overall generalization is not systematic, as evidenced by high variance across different \suppcomp, especially for higher complexity and more novel compositions.}
\end{figure*}

Figure \ref{fig:heatmaps} breaks down the performance of T5 on \mix{$T_{12}$} after training on \diverse{$T_{12}$}. The heatmaps plot performance across various \reasonpat (\suppcomp) occurring in the test data, sub-divided by the number of required supporting facts $n$ per question. Performance on support compositions seen at training time (blue frames) is generally high, indicating the importance of training pattern diversity for better generalization.
The plots indicate that T5 shows some ability to generalize to new support compositions, especially for lower $n$. Furthermore, certain question types appear to be more learned more robustly; for \listq and \countq questions, performance remains relatively high even for larger $n$ and across novel \suppcomp. We hypothesize that such questions may be easier as simple counting rules suffice to reach an answer, and these are ``close to the surface''; unlike other events that may implicitly convey information, in our stories, changes of possession are always explicit in the text. 

In general however, the plots indicate that T5 is far from robust compositional generalization:

\xhdr{Performance deteriorates with increased complexity} Performance is near perfect for simple compositions ($n \leq 2$) but deteriorates significantly for more complex cases (e.g., center and right plots).

\xhdr{Inconsistent knowledge} The discrepancy between the relatively high performance on \wherep questions compared with very low performance on \yesno questions suggests that models aren't learning consistent knowledge representations. E.g., if a model answers $y$ correctly to some ``Where is $p$?'' question, we would expect it to answer ``yes'' correctly for the same question in \yesno format, ``Is $p$ at $y$?''. We present further empirical support for this finding in appendix \ref{ssec:ext_err_analysis_know_incon}.

\xhdr{Performance below chance for certain question types} The heatmaps expose a particularly challenging class of \yesno questions involving disjunctions over indefinites (center and right plots, bottom right); accuracy for such questions is close to zero. See appendix \ref{ssec:ext_err_double_disj} for an example instance.

\section{Future work \& conclusions}
\label{sec:fw_conclusions}
Our work opens up multiple new directions for future research. \pybabi is readily extendable, for systematic probing of more diverse linguistic phenomena. A beneficial first step could include integration of other bAbI tasks such as spatial reasoning and agent motivations. That said, our experience suggests that the design of truly scalable synthetic and dynamic benchmarks poses significant theoretical and engineering challenges, warranting deeper research on their own right.

Our results raise new questions about the viability of learning robust situation models using standard question-answering training methods, and our datasets present new modelling challenges for future efforts. 

Additionally, \pybabi can naturally complement parallel work probing the the situation representations constructed by neural language models~\citep{li-etal-2021-implicit}, by facilitating tailored data generation for specific questions, thus broadening and deepening the scope of possible research.

In conclusion, we introduced \pybabi, a new framework for highly controllable bAbI task generation. We used it to create compositional generalization datasets providing new modelling challenges for state-of-the-art neural language models. More broadly, our results underscore the importance of agile development of benchmarks themselves, beyond only the models solving them.

\section*{Acknowledgements}
We thank the Aristo team at the Allen Institute for AI for valuable support and feedback at various stages of this work. Ronen Tamari was supported by the Center for Interdisciplinary Data-science Research (CIDR) at HUJI. This work was supported by the European Research Council (ERC) under the European Union's Horizon 2020 research and innovation programme (grant no. 852686, SIAM) and NSF-BSF grant no. 2017741 (Shahaf). Part of this research is also supported by   the European Research Council, ERC-StG grant no.\ 677352 (Tsarfaty), which we gratefully acknowledge.  
\bibliography{anthology,custom}
\bibliographystyle{acl_natbib}

\clearpage
\newpage

\appendix

\section{Appendix}
\label{sec:appendix}
\subsection{Extended task construction details}
\label{ssec:ext_task_details}

This section provides further details of the training and test splits used for our experiments. 

Table \ref{tab:concepts} enumerates the basic ``building blocks'', or concepts underlying the tasks, as presented in \S\ref{ssec:task_structure}.

Tables \ref{tab:app_cid_7} and \ref{tab:app_cid_12} detail the concept sets for each of the sub-tasks comprising the training and test sets, for the $T_2$, $T_7$ and $T_{12}$ groups of tasks.

\begin{table*}[]
\small
\begin{tabular}{@{}llll@{}}
\toprule
Events                                                           & Template                                                                                                                                                     & Example                                                                                                       & Notes                                                                                             \\ \midrule
MOVE                                                             & P \{moved\} to the L.                                                                                                                                        & John traveled to the park.                                                                                    &                                                                                                   \\
GRAB                                                             & P \{grabbed\} the O.                                                                                                                                         & Mary picked up the apple.                                                                                     &                                                                                                   \\
DROP                                                             & P \{dropped\} the O.                                                                                                                                         & Daniel dropped the milk.                                                                                      &                                                                                                   \\
GIVE                                                             & P1 \{gave\} P2 the O.                                                                                                                                        & John handed Mary the apple.                                                                                   &                                                                                                   \\ \midrule
\begin{tabular}[c]{@{}l@{}}Linguistic \\ Constructs\end{tabular} &                                                                                                                                                              &                                                                                                               &                                                                                                   \\ \midrule
COREF                                                            & \begin{tabular}[c]{@{}l@{}}P (MOVE|GRAB|DROP)\\ Following that, \{he\} (MOVE|GRAB|DROP).\end{tabular}                                                        & \begin{tabular}[c]{@{}l@{}}John went to the garden.\\ Following that, he moved to the store\end{tabular}      & Co-reference                                                                                      \\
CONJ                                                             & P1 and P2 \{moved\} to the L1.                                                                                                                               & Jeff and Fred went to the cinema.                                                                             & Conjunction                                                                                       \\
COMPOUND                                                         & \begin{tabular}[c]{@{}l@{}}P1 and P2 \{moved\} to the L1.\\ Then they \{moved\} to the L2.\end{tabular}                                                      & \begin{tabular}[c]{@{}l@{}}Jeff and Fred went to the cinema.\\ Then they traveled to the school.\end{tabular} & Compound co-reference                                                                             \\
NEGATE                                                           & P is not at the L.                                                                                                                                           & Julie is not in the park.                                                                                     & Negation                                                                                          \\
INDEF                                                            & P is either at the L1 or the L2.                                                                                                                             & John is either in the park or the school.                                                                     & Indefinite expression                                                                             \\ \midrule
Questions                                                        &                                                                                                                                                              &                                                                                                               &                                                                                                   \\ \midrule
where-P                                                          & Where is P?                                                                                                                                                  & Where is John?                                                                                                &                                                                                                   \\
where-O                                                          & Where is the O?                                                                                                                                              & Where is the football?                                                                                        &                                                                                                   \\
where-was-O                                                      & Where was the O before the L?                                                                                                                                & Where was the football before the hallway?                                                                    &                                                                                                   \\
yes-no                                                           & Is P at the L?                                                                                                                                               & Is John at the park?                                                                                          &                                                                                                   \\
list                                                             & What is P carrying?                                                                                                                                          & What is John carrying?                                                                                        &                                                                                                   \\
counting                                                         & How many objects is P carrying?                                                                                                                              & How many objects is John carrying?                                                                            &                                                                                                   \\ 
give-qs                                                          & \begin{tabular}[c]{@{}l@{}}Who gave the O to P2?\\ Who gave the O?\\ Who received the O?\\ Who did P1 give the P2 to?\\ What did P1 give to P2?\end{tabular} & \begin{tabular}[c]{@{}l@{}}Who gave the football to John?\\ ...\end{tabular}                                  & \begin{tabular}[c]{@{}l@{}}Constitutes multiple\\ question types over GIVE\\ events.\end{tabular} \\ \bottomrule
\end{tabular}
\caption{\label{tab:concepts} Details of the events, linguistic constructs and questions constituting the bAbI tasks covered in this work. Words in \{brackets\} are drawn from a small set of synonyms.}
\end{table*}

As can be seen from the tables, the main sources of compositionality are:
\begin{compactitem}
    \item Following the \babibm task structure, at training time, all of the more complex linguistic constructs are seen only with MOVE events (and none of the other event types).
    \item Similarly, at training time, \yesno questions are always seen only with MOVE events (and none of the other event types), and with the INDEF or NEGATE linguistic constructs (but not others, such as COREF).
    \item \wherewaso questions are never seen in stories with GIVE events.
\end{compactitem}

\subsubsection{Example instances}
\begin{figure*}[]
\centering
\includegraphics[width=\linewidth]{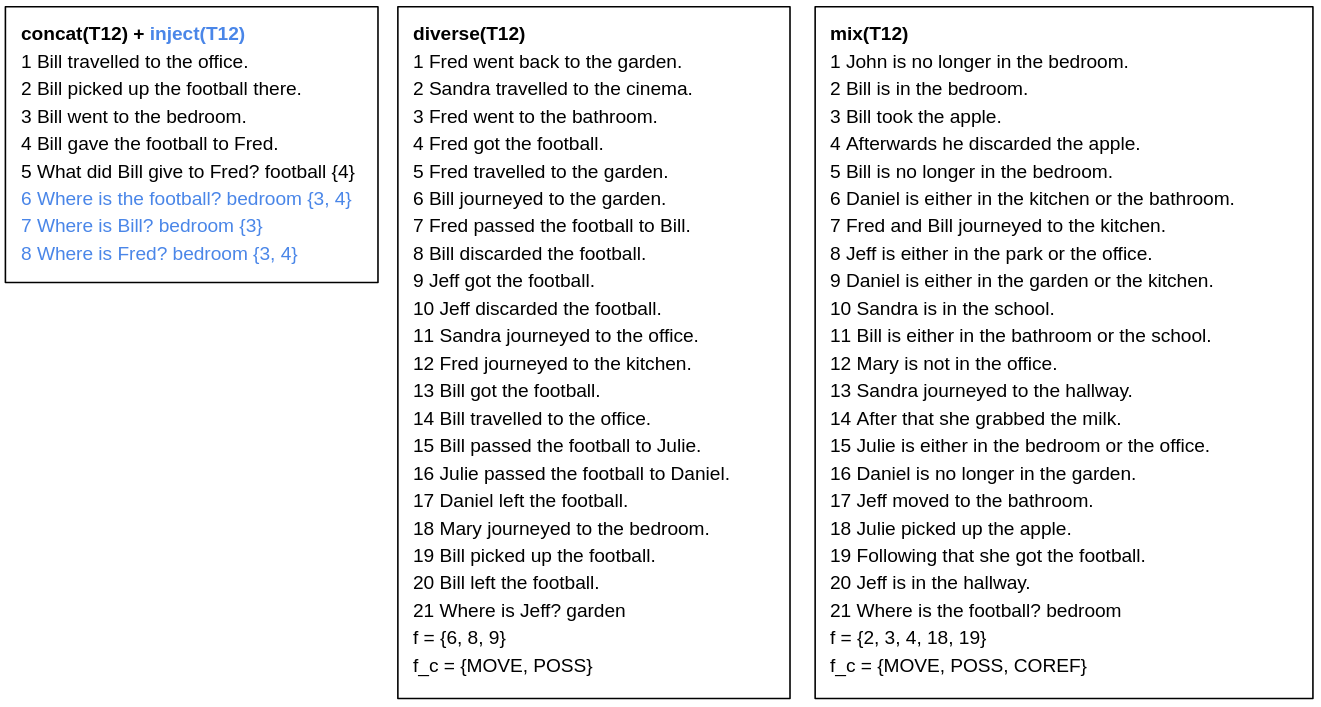}

\caption{\label{fig:split_examples} Example instances from each of the 4 types of splits used in our experiments.}
\end{figure*}

Figure \ref{fig:split_examples} shows examples from each of the 4 types of splits used in our experiments. The \concatnarg instance is from the original \babibm task 5. The \injectnarg data contains the same passages as \concatnarg, but adds supplementary questions on agent and object locations. \diversenarg instances contain more diverse support compositions (\suppcomp), but certain combinations are held out. In particular, \diversenarg instances only feature non-default linguistic mappings with MOVE events, never with POSS (GRAB or DROP) or GIVE. In the \mixnarg instances, all combinations of support compositions are possible, as shown in the example which features possession (POSS) events along with co-reference.

\begin{table*}[]
\small
\centering 
\begin{tabular}{@{}lllllllllllll@{}}
\toprule
         &       & \multicolumn{4}{c}{Events}                                                                                                               & \multicolumn{3}{c}{Linguistic Constructs}                                                                                 & \multicolumn{4}{c}{Questions}                                                                                                      \\ \midrule
Sub-task & Type  & \rot{Move} & \rot{Grab} & \rot{Drop} & \rot{Give}                    & \rot{Co-reference} & \rot{Conjunction} & \rot{Compound co-ref.}        & \rot{where-P} & \rot{where-O} & \rot{where-was-O} & \rot{give} \\ \midrule
1        & Train & \Checkmark   &                             &                             & \multicolumn{1}{l|}{}                          &                                     &                                    & \multicolumn{1}{l|}{}                          & \Checkmark      &                                &                                    &                             \\
2        & Train & \Checkmark   & \Checkmark   & \Checkmark   & \multicolumn{1}{l|}{}                          &                                     &                                    & \multicolumn{1}{l|}{}                          & I/D                            & \Checkmark      &                                    &                             \\
3        & Train & \Checkmark   & \Checkmark   & \Checkmark   & \multicolumn{1}{l|}{}                          &                                     &                                    & \multicolumn{1}{l|}{}                          & I                              & I                              & \Checkmark          &                             \\
5        & Train & \Checkmark   & \Checkmark   & \Checkmark   & \multicolumn{1}{l|}{\Checkmark} &                                     &                                    & \multicolumn{1}{l|}{}                          & I/D                            & I/D                            &                                    & \Checkmark   \\
11       & Train & \Checkmark   &                             &                             & \multicolumn{1}{l|}{}                          & \Checkmark           &                                    & \multicolumn{1}{l|}{}                          & \Checkmark      &                                &                                    &                             \\
12       & Train & \Checkmark   &                             &                             & \multicolumn{1}{l|}{}                          &                                     & \Checkmark          & \multicolumn{1}{l|}{}                          & \Checkmark      &                                &                                    &                             \\
13       & Train & \Checkmark   &                             &                             & \multicolumn{1}{l|}{}                          &                                     &                                    & \multicolumn{1}{l|}{\Checkmark} & \Checkmark      &                                &                                    &                             \\ \midrule

mix($T_{2}$)  & Test  & \Checkmark   & \Checkmark   & \Checkmark   & \multicolumn{1}{l|}{} & \Checkmark           &           & \multicolumn{1}{l|}{} &       & \Checkmark      &           &                             \\

mix($T_{7}$)  & Test  & \Checkmark   & \Checkmark   & \Checkmark   & \multicolumn{1}{l|}{\Checkmark} & \Checkmark           & \Checkmark          & \multicolumn{1}{l|}{\Checkmark} & \Checkmark      & \Checkmark      & \Checkmark          &                             \\ \bottomrule
\end{tabular}
\caption{\label{tab:app_cid_7} Concept sets for the $T_{2}$ and $T_{7}$ sub-set of the original bAbI tasks, and the new tasks generated with \pybabi. Train sub-task numbering follows the original bAbI numbering. The \injectnarg and \diversenarg tasks inherit the same concept set from the original tasks, and additionally ``I'', ``D'' denote question types included only in the \injectnarg or \diversenarg tasks, respectively. ``I/D'' denotes question types included in both.}
\end{table*}

\begin{table*}[]
\small
\begin{tabular}{@{}llllllllllllllllll@{}}
\toprule
         &       & \multicolumn{4}{l}{Events}                                                                                                               & \multicolumn{4}{l}{Linguistic Constructs}                                                                                                                 &                                                & \multicolumn{7}{l}{Questions}                                                                                                                                                                                                      \\ \midrule
Task     & Type  & \rot{Move} & \rot{Grab} & \rot{Drop} & \rot{Give}                    & \rot{Co-reference} & \rot{Conjunction} & \rot{Compound co-ref.} & \rot{Negation} & \rot{Indefinite}              & \rot{where-P} & \rot{where-O} & \rot{where-was-O} & \rot{yes-no} & \rot{counting} & \rot{list} & \rot{give} \\ \midrule
1        & Train & \Checkmark   &                             &                             & \multicolumn{1}{l|}{}                          &                                     &                                    &                                              &                                 & \multicolumn{1}{l|}{}                          & \Checkmark      &                                &                                    &                               &                                 &                             &                             \\
2        & Train & \Checkmark   & \Checkmark   & \Checkmark   & \multicolumn{1}{l|}{}                          &                                     &                                    &                                              &                                 & \multicolumn{1}{l|}{}                          & I/D                            & \Checkmark      &                                    &                               &                                 &                             &                             \\
3        & Train & \Checkmark   & \Checkmark   & \Checkmark   & \multicolumn{1}{l|}{}                          &                                     &                                    &                                              &                                 & \multicolumn{1}{l|}{}                          & I                              & I                              & \Checkmark          &                               &                                 &                             &                             \\
5        & Train & \Checkmark   & \Checkmark   & \Checkmark   & \multicolumn{1}{l|}{\Checkmark} &                                     &                                    &                                              &                                 & \multicolumn{1}{l|}{}                          & I/D                            & I/D                            &                                    &                               &                                 &                             & \Checkmark   \\
6        & Train & \Checkmark   &                             &                             & \multicolumn{1}{l|}{}                          &                                     &                                    &                                              &                                 & \multicolumn{1}{l|}{}                          & I/D                            &                                &                                    & \Checkmark     &                                 &                             &                             \\
7        & Train & \Checkmark   & \Checkmark   & \Checkmark   & \multicolumn{1}{l|}{\Checkmark} &                                     &                                    &                                              &                                 & \multicolumn{1}{l|}{}                          & I                              & I                              &                                    &                               & \Checkmark       &                             &                             \\
8        & Train & \Checkmark   & \Checkmark   & \Checkmark   & \multicolumn{1}{l|}{}                          &                                     &                                    &                                              &                                 & \multicolumn{1}{l|}{}                          & I                              & I                              &                                    &                               &                                 & \Checkmark   &                             \\
9        & Train & \Checkmark   &                             &                             & \multicolumn{1}{l|}{}                          &                                     &                                    &                                              & \Checkmark       & \multicolumn{1}{l|}{}                          & I/D                            &                                &                                    & \Checkmark     &                                 &                             &                             \\
10       & Train & \Checkmark   &                             &                             & \multicolumn{1}{l|}{}                          &                                     &                                    &                                              &                                 & \multicolumn{1}{l|}{\Checkmark} & I/D                            &                                &                                    & \Checkmark     &                                 &                             &                             \\
11       & Train & \Checkmark   &                             &                             & \multicolumn{1}{l|}{}                          & \Checkmark           &                                    &                                              &                                 & \multicolumn{1}{l|}{}                          & \Checkmark      &                                &                                    &                               &                                 &                             &                             \\
12       & Train & \Checkmark   &                             &                             & \multicolumn{1}{l|}{}                          &                                     & \Checkmark          &                                              &                                 & \multicolumn{1}{l|}{}                          & \Checkmark      &                                &                                    &                               &                                 &                             &                             \\
13       & Train & \Checkmark   &                             &                             & \multicolumn{1}{l|}{}                          &                                     &                                    & \Checkmark                    &                                 & \multicolumn{1}{l|}{}                          & \Checkmark      &                                &                                    &                               &                                 &                             &                             \\ \midrule
mix({$T_{12}$}) & Test  & \Checkmark   & \Checkmark   & \Checkmark   & \multicolumn{1}{l|}{\Checkmark} & \Checkmark           & \Checkmark          & \Checkmark                    & \Checkmark       & \multicolumn{1}{l|}{\Checkmark} & \Checkmark      & \Checkmark      & \Checkmark          & \Checkmark     & \Checkmark       & \Checkmark   &                             \\ \bottomrule
\end{tabular}
\caption{\label{tab:app_cid_12} Concept sets for the $T_{12}$ sub-set of the original bAbI tasks, and the new tasks generated with \pybabi. Train sub-task numbering follows the original bAbI numbering. The \injectnarg and \diversenarg tasks inherit the same concept set from the original tasks, and additionally ``I'', ``D'' denote question types included only in the \injectnarg or \diversenarg tasks, respectively. ``I/D'' denotes question types included in both.}
\end{table*}

\subsubsection{Long instances in the \babibm tasks}
For the T5 experiments, we used a slightly modified version of the \babibm tasks, where we trimmed all training and validation examples that didn't fit into the 512-token input window. This resulted in trimming 1,585 training instances and 175 validation instances from $T_7$ and $T_12$ (common to both sets). These data points are not consequential as our analysis focuses on the effects of compositionality and not story length; all instances in \diversenarg and \mixnarg are substantially shorter than the 512-token maximum input window size.

\subsection{Implementation details}
\label{ssec:impl}

\xhdr{T5} We use the publicly available HuggingFace pre-trained T5-base implementation~\citep{wolf-etal-2020-transformers}. We fine-tune T5 for 12 epochs on our bAbI data, using the Adam optimizer~\citep{kingma2017adam}, an initial learning rate of $5 * 10^{-5}$ and training batch size of 8.

\xhdr{STM} We used the official STM implementation\footnote{\url{https://github.com/thaihungle/SAM}}, with the only change being a batch size of 32 instead of 128, due to technical constraints.

\xhdr{EntNet} We re-implemented the model in PyTorch, similarly using a batch-size of 32. Following the official Lua reference implementation\footnote{\url{https://github.com/facebookarchive/MemNN/tree/master/EntNet-babi}}, we used 20 memory units each with dimension 100. We used the SGD optimizer.

For both the EntNet and STM, we trained models for 200 epochs, and took the best of 10 tries, following ~\citet{entnet2017}.

For the 20-model concurrence benchmark, refer to \citet{liu2021small} for model details, as we used the same experimental setup.

For the T5 experiments, we used the PyTorch Lightning~\citep{falcon2019pytorch} trainer implementation, and Weights \& Biases~\citep{wandb} for experiment tracking and artifacts management.

\subsection{Inoculation experiment results}
\label{ssec:ext_inoc}
\begin{figure}
\centering
\includegraphics[width=\columnwidth]{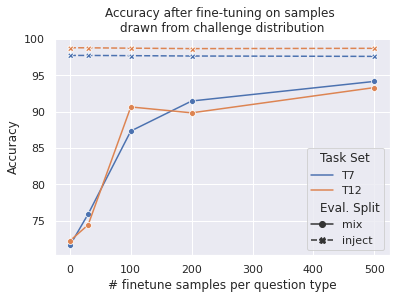}

\caption{\label{fig:inoc} Inoculation experiment results.}
\end{figure}

 To rule out the hypothesis that certain patterns may be too hard for models to learn, we follow the inoculation methodology presented in \citet{liu-etal-2019-inoculation}: after training on the original tasks, we fine-tune the T5 on small amounts of OOD data (disjoint from the test data), and evaluate performance as a function of ``inoculation dose''. As can be seen in Fig. \ref{fig:inoc}, we find that performance quickly (with only 500 additional inoculation samples per question type) reaches over 90\% accuracy on both the \mix{$T_{7}$} and \mix{$T_{12}$} challenge sets. These results support the hypothesis that the training data is not rich enough, indicating clearly that the model is capable of quickly learning to solve the challenge tasks, given exposure to training samples with similar enough patterns.

\subsection{Concurrence experiments}
\label{ssec:ext_concurrence}
Table \ref{tab:concurrence_full} presents the full results for the concurrence experiments of \S\ref{ssec:4_1_analysis}. SQuAD and bAbI task 2 results are reproduced from \citet{liu2021small}, see there also for implementation details of the models used.

\begin{table}[]
\small
\begin{tabular}{@{}l@{\hspace{1.2\tabcolsep}}l@{\hspace{1.2\tabcolsep}}l@{\hspace{1.2\tabcolsep}}l@{\hspace{1.2\tabcolsep}}l@{}}
\toprule
Model                                                                   & \multicolumn{4}{c}{Evaluation accuracy} \\ \midrule
                                                                        & SQuAD & mix(T2) & mix(T7) & babi task 2 \\
rasor                                                                   & 64.86 & 88.20   & 35.03   & 100.00      \\
bidaf                                                                   & 67.39 & 97.20   & 30.50   & 100.00      \\
documentreader                                                          & 69.66 & 90.20   & 40.70   & 100.00      \\
\begin{tabular}[c]{@{}l@{}}documentreader\\ (no\_features)\end{tabular} & 69.21 & 82.50   & 37.17   & 100.00      \\
bidafplusplus                                                           & 69.49 & 99.50   & 44.20   & 80.70       \\
mnemonicreader                                                          & 73.02 & 98.20   & 39.63   & 100.00      \\
\begin{tabular}[c]{@{}l@{}}mnemonicreader\\ (no\_features)\end{tabular} & 72.67 & 97.50   & 38.20   & 100.00      \\
qanet                                                                   & 72.41 & 67.70   & -       & 100.00      \\
fusionnet                                                               & 72.90 & 99.50   & 39.73   & 100.00      \\
\begin{tabular}[c]{@{}l@{}}fusionnet\\ (no\_features)\end{tabular}      & 72.24 & 88.10   & 37.80   & 100.00      \\
bert                                                                    & 81.46 & 95.50   & 47.63   & 100.00      \\
bert\_large                                                             & 84.17 & 98.30   & 59.10   & 100.00      \\
bert\_large\_wwm                                                        & 87.33 & 98.70   & 67.63   & 99.90       \\
albert                                                                  & 81.86 & 98.20   & 56.70   & 100.00      \\
albert\_xxlarge                                                         & 89.07 & 99.80   & 80.00   & 100.00      \\
roberta                                                                 & 83.37 & 98.70   & 57.70   & 100.00      \\
roberta\_large                                                          & 86.96 & 99.80   & 64.07   & 100.00      \\
electra                                                                 & 85.88 & 98.70   & 53.47   & 100.00      \\
spanbert                                                                & 86.20 & 98.40   & 55.70   & 99.50       \\
spanbert\_large                                                         & 88.74 & 98.60   & 62.27   & 95.40       \\ \bottomrule
\end{tabular}
\caption{\label{tab:concurrence_full} Full results of concurrence experiments presented in \S\ref{ssec:4_1_analysis}.}
\end{table}

\subsection{Extended error analysis: GIVE events}
\label{ssec:ext_err_analysis_give}
We analyze the performance of models on the \mix{$T_7$} split after being trained on \concat{$T_7$}, and in particular we focus on GIVE events. As noted in \S\ref{ssec:exp_2}, compositions involving GIVE are intuitively challenging as they entail multiple inferences which are not explicit in the text: the actors share the same location, and the possession of the object being given is transferred from the giver to the recipient. The only task in \concat{$T_7$} featuring GIVE events is task 5, which never asks about the locations of actors or objects, but only about the participant roles in the event (e.g., who was the giver or recipient; see Fig. \ref{fig:sdb-overview} example from task 5).

\begin{table}[]
\small
\begin{tabular}{@{}lllll@{}}
\toprule
\# supporting facts & \# samples & \multicolumn{3}{l}{Evaluation accuracy} \\ \midrule
                    &            & BiDAF       & RoBERTa      & T5         \\
1                   & 334        & 53.3        & 93.4         & 86.8       \\
2                   & 833        & 45.7        & 73.3         & 72.4       \\
-- (including give)  & 99         & 3.03        & 7.07         & 12.12      \\
3                   & 1832       & 19.4        & 37           & 53.8       \\
-- (including give)  & 468        & 4.27        & 7.05         & 30.3       \\ \bottomrule
\end{tabular}
\caption{\label{tab:exp_1_breakdown} Breakdown of model performance on \mix{$T_7$} for questions including GIVE events in the supporting fact set. The poor performance indicates that training on the \babibm data does not facilitate generalization to novel compositions of GIVE.}
\end{table}

To measure this intuition empirically, we analyze a subset of 567 questions including GIVE events in the supporting facts set. As shown in Table \ref{tab:exp_1_breakdown}, performance for all models on questions including GIVE is extremely low, far below performance for questions without it. Qualitative analysis indicates many failure cases follow the pattern shown in the right-side example of Fig. \ref{fig:sdb-overview}c, question on line 10: the location of an entity (e.g., Daniel) must be inferred via the known (co-)location of a second participant in the GIVE event (e.g., Jeff). These results strengthen the hypothesis that standard QA training on the original bAbI data does not drive strong event comprehension in models.

\subsection{Extended error analysis: knowledge inconsistency}
\label{ssec:ext_err_analysis_know_incon}
This section presents further analysis of the knowledge consistency of the T5 model trained on the \diverse{$T_{12}$} data when evaluated on the challenge set \mix{$T_{12}$}.

We collected all \yesno questions from \mix{$T_{12}$} for which the answer was ``yes'', yielding 446 questions in total. For each such (question, answer) pair, of the form (``Is \texttt{person} at the \texttt{location}?'', ``yes''), we created an equivalent pair in the format of a \wherep question, (``Where is \texttt{person}?'', \texttt{location}).

\begin{figure}[]
\centering
\includegraphics[width=\columnwidth]{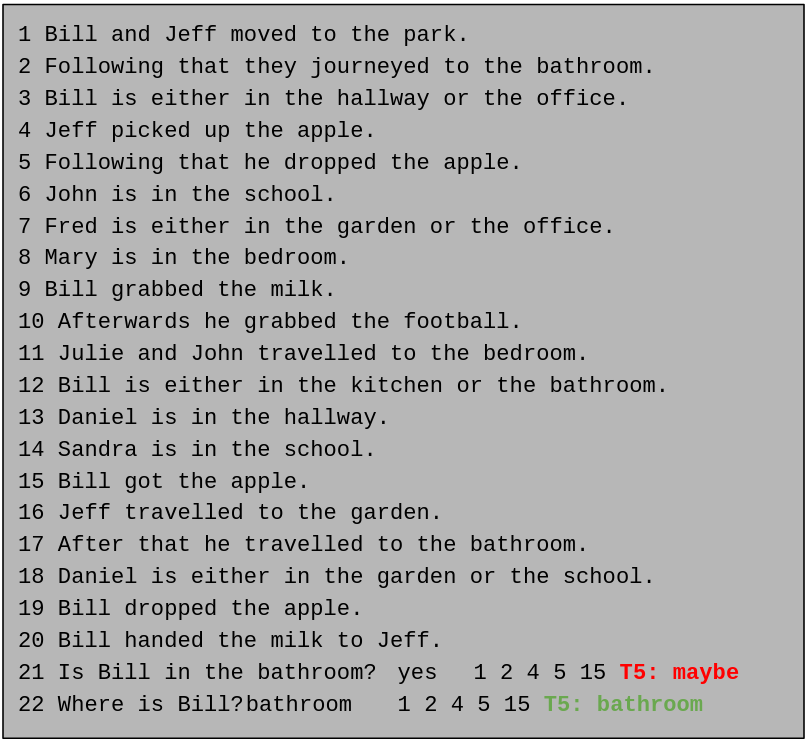}

\caption{\label{fig:kn_con_example} Example \mix{$T_{12}$} instance displaying model knowledge inconsistency: T5 correctly answers the question in \wherep form (line 22), and incorrectly in \yesno form (line 21).}
\end{figure}

Ideally, we would expect a model to be agnostic to equivalent phrasings of a question. However, as displayed in Figure \ref{fig:kn_consist}, we find that T5 is considerably more accurate for questions posed in the \wherep format, likely due to exposure to a larger variety of such questions at training time. Figure \ref{fig:kn_con_example} shows a characteristic example: T5 correctly answers in the \wherep format, but wrongly answers ``maybe'' for the \yesno format, thrown off by the distractor indefinite phrase in sentence 3. The pattern of answering ``maybe'' to questions about the location of an actor mentioned in an indefinite is commonly observed in training.

\begin{table}[]
\centering
\begin{tabular}{|l|l|l|}
\hline
\begin{tabular}[c]{@{}l@{}}\wherep ($\rightarrow$)\\ \yesno ($\downarrow$)\end{tabular} & correct   & incorrect  \\ \hline
correct                                                          & 209 & 4  \\ \hline
incorrect                                                          & 145 & 88 \\ \hline
\end{tabular}
\caption{\label{fig:kn_consist}  Confusion matrix displaying knowledge inconsistencies in T5. We pose a question in two formats: (1) \yesno: ``Is \textit{X} at \textit{L}? yes'' vs (2) \wherep: ``Where is \textit{X}? \textit{L}''. We find performance is considerably higher for questions posed in the \wherep format, indicating the model isn't learning the equivalence of both forms.}
\end{table}

These results highlight a limitation of text-to-text QA models such as T5: their story representation may be highly coupled with the input question. This form of representation stands in contrast to more human-like narrative comprehension which is thought to involve the construction of situation models;  structured representations of entities and their relations as depicted by the text. Situation models are not dependent on a-priori knowledge of a particular question, and are thought to constitute a representational substrate supporting more systematic inferential generalization.

\subsection{Extended error analysis: double disjunctions}
\label{ssec:ext_err_double_disj}

As the shown in the \S\ref{ssec:exp_2_err_analysis} error analysis, a particularly difficult class of questions are double disjunctions over indefinite expressions. Figure \ref{fig:double_disj_1} displays a typical example from \mix{$T_{12}$}, where the locations of two actors are given in indefinite form (sentences 3 and 19), and are also known to be co-located, since they share the location of the object ``football'', as inferred from sentences 18 and 20. Hence it is possible to infer their location as the intersection of the two indefinite expressions (here ``bedroom''). Rather than answering ``yes'' to the question ``Is John in the bedroom?'', T5 invariably answers ``maybe'' for such cases. This pattern is likely due to the fact that in the training data ``maybe'' is a typical answer for \yesno questions about actors mentioned by indefinite expressions (task 10 in \babibm).

\begin{figure}[!t]
\centering 
\fbox{\begin{minipage}{15em}
\small
1 Bill grabbed the milk.\\
2 Bill put down the milk.\\
3 John is either in the bedroom or the kitchen.\\
4 Fred journeyed to the kitchen.\\
5 John grabbed the football.\\
6 Following that he put down the football.\\
7 Bill picked up the milk.\\
8 Following that he went to the bedroom.\\
9 Bill is in the office.\\
10 Bill is in the cinema.\\
11 Bill passed the milk to Julie.\\
12 Julie handed the milk to Bill.\\
13 Jeff is not in the school.\\
14 John took the football.\\
15 Fred and Jeff moved to the school.\\
16 Afterwards they journeyed to the bathroom.\\
17 Bill handed the milk to Julie.\\
18 John dropped the football.\\
19 Daniel is either in the school or the bedroom.\\
20 Daniel took the football.\\
21 Is John in the bedroom?	yes	3 18 19 20
\end{minipage}
}
\caption{Double disjunction example from \mix{$T_{12}$}.}
\label{fig:double_disj_1}
\end{figure}

\end{document}